%% file: main.tex
\documentclass{article}

\usepackage[preprint]{neurips_2022}
\usepackage[utf8]{inputenc} 
\usepackage[T1]{fontenc}    
\usepackage{hyperref}       
\usepackage{url}            
\usepackage{booktabs}       
\usepackage{amsfonts}       
\usepackage{nicefrac}       
\usepackage{microtype}      
\usepackage{xcolor}         
\usepackage{graphicx}
\usepackage{subcaption}
\usepackage{caption}
\usepackage{amsmath}
\usepackage{array}

\title{Conditional Latent ODEs for Motion Prediction in Autonomous Driving}

\author{%
  Truong G.~Khang \\
  KAIST CS \\ 
  \texttt{khangtg@kaist.ac.kr} \\
  \And
  Yongjae~Kim\\
  KAIST ME \\ 
  \texttt{kim4375731@kaist.ac.kr} \\
  \And
  Andrea~Finazzi\\
  The Robotics Program - KAIST EE\\
  \texttt{finazzi@kaist.ac.kr} \\
}

\begin{document}

\maketitle

\input{sections/00_abstract}
\input{sections/10_introduction}
\input{sections/20_related_works} 
\input{sections/30_methodology}
\input{sections/40_results}
\input{sections/50_discussion_and_conclusions}

\bibliography{ref}

\input{sections/60_appendix}

\end{document}

%% file: sections/00_abstract.tex
\begin{abstract}
  This paper addresses imitation learning for motion prediction problem in autonomous driving, especially in multi-agent setting. Different from previous methods based on GAN, we present the conditional latent ordinary differential equation (cLODE) to leverage both the generative strength of conditional VAE and the continuous representation of neural ODE. Our network architecture is inspired from the Latent-ODE model. The experiment shows that our method outperform the baseline methods in the simulation of multi-agent driving and is very efficient in term of GPU memory consumption. Our code and docker image are publicly available: \verb+https://github.com/TruongKhang/cLODE+; \verb+https://hub.docker.com/r/kim4375731/clode+.
\end{abstract}

%% file: sections/10_introduction.tex
\section{Introduction} \label{intro}




In the field of autonomous driving, predicting the motion of the surrounding agents is a crucial and challenging task to solve complex navigation problems in dynamic environments. Imitation Learning (IL) is a set of well-known approaches that try to model the behaviour of drivers from real-world data. Behavioural Cloning (BC), Inverse Reinforcement Learning (IRL), and Generative Adversarial IL (GAIL) are three of the most popular approaches. They formulate the imitation task with a supervised learning approach, as a policy optimization problem, or via a GAN-inspired generative procedure, respectively.

However, a main drawback of the above-mentioned methods is that the generated trajectories cannot capture time-wise correlation. In particular, relying on RNNs, these models learn a transformation between subsequent observations, failing at representing consistency of purpose in the generated driving behaviour.

Motivated by this analysis, we propose to control the generated trajectories by using a Neural Ordinary Differential Equation (NODE) \citep{chen2018neuralode}. To build a generative model for imitation learning, we use a Variational Auto-Encoder (VAE) which is easier to train with ODE compared to GAN. We achieve these two goals by using Latent-ODE \citep{rubanova2019lode} which can explicitly learn a representation of the trajectory space. The generated trajectories sampled from this space mimic the driving style of the expert demonstrations. Our method takes the history data, including the actions and the observations in the past, to estimate a posterior for generating the action trajectory. Therefore, we regard our method as a conditional variation of Latent ODEs (cLODE).

The contribution of this work is twofold:
\begin{itemize}
    \item We propose a conditional Latent-ODE model for imitation learning.
    \item We train and evaluate our method by using the US-101 dataset \citep{colyar2007us} in a multi-agent setting. The experiments demonstrate that our method outperforms the baseline AGen \citep{si2019agen}.
\end{itemize}





%% file: sections/20_related_works.tex
\section{Related works} \label{related}



\paragraph{Imitation learning} Imitation learning \citep{hussein2017reviewimitation} is a set of techniques aiming at mimicking the behaviour of an expert agent by learning from data or, as this approach is often referred to, learning from demonstration. Among the most popular algorithms, adopted in many works also in the context of autonomous driving to solve diverse tasks, are Behavioral Cloning (BC), Inverse Reinforcement Learning (IRL), and Generative Adversarial Imitation Learning (GAIL). 

BC \citep{bain1995bc}, formulated as a supervised learning approach, is a data-efficient approach that has the advantage of being relatively easy to implement \citep{kuefler2017gail-driving}. However, as it relies on the assumption that the data is i.i.d., BC is particularly sensible to the issue known as covariate shift \citep{kuefler2017gail-driving}.


To cope with this limitation, IRL \citep{ho2016irl, finn2016irl, ho2016gail} solves the imitation learning problem by learning the unknown reward function which the expert is supposed to be acting upon. Although this method well generalizes to out-of-distribution situations, its dual-step nature makes it less efficient than direct learning approaches, such as GAIL.


Generative adversarial reinforcement learning, deriving from Generative Adversarial Networks (GANs), trains a generator model to fool a critic to the point that the latter is unable to distinguish between generated and expert trajectories \citep{ho2016gail, kuefler2017gail-driving}. \citet{song2018ma-gail, bhattacharyya2018ps-gail} introduce multi-agent GAIL, which has been extended by \citet{si2019agen} with recursive online adaptation.

Also, several works combine one or more of these solutions to overcome well-known limitations \citep{jena2021augmentinggail, sasaki2020abc, wang2019ilhpp, torabi2018bco}.

\paragraph{Latent ordinary differential equations}
Recurrent and residual models, as well as normalizing flows, perform the generative process by learning a transformation that is applied in an iterative manner to the latent variable. 

\citet{chen2018neuralode} define the evolution of the hidden variable in the time limit, parameterizing its dynamics with neural network-defined ordinary differential equations. The resulting representation describes the nature of the latent space in the continuous domain, offering several advantages in terms of memory and parameter efficiency, scalability and flexibility with respect to time-wise constraints. 

As an evolution of this work, \citet{rubanova2019lode} define a continuous-time state dynamics model for RNNs, combining the power of recurrent networks for structured data generation with the highly information-efficient ODE-based latent dynamics introduced by \citet{chen2018neuralode}. The models introduced by the authors (ODE-RNN and Latent ODE) reduce data loss and are better suited for irregularly-sampled data, compared to other works aiming at improving RNN time flexibility \citep{che2018rnndecay, lipton2016rnndecay}.


%% file: sections/30_methodology.tex
\section{Method}  \label{method}

\subsection{Conditional Latent ODEs for imitation learning}
Let $\tau$ be an expert trajectory which consists of $T$ observations and actions $\tau=(a_1,o_1,…,a_T,o_T )$. We denote $h_{t-1}=(a_1,o_1,…,a_{t-1},o_{t-1})$ as the history of actions and observations before the time step $t$. The goal of imitation learning is to learn a policy $\pi_\theta = p(a_t | h_{t-1}; \theta)$,  parameterized by $\theta$, which can imitate the expert behaviour $\tau$ and predict an action at any time step $t$ from the history of actions and observations $h_{t-1}$. 

We estimate $\theta$ by maximizing the likelihood with respect to the trajectory $\tau$, $p(\tau | \theta)$. Because each observation $o_t$ can be estimated from the action $a_t$ in a deterministic way via a kinematic model, the likelihood can be regarded as the likelihood of the action variables only. Therefore, the likelihood $p(\tau | \theta)$ is represented via the action distribution as follows

\begin{equation}
    \label{likelihood_tau}
    p(\tau | \theta) = p(a_1 | \theta) \prod_{t=1}^{T-1} p(a_{t+1}|h_t,\theta)
\end{equation}

The action distribution conditioned to the history data is the policy that we aim to learn. To compute $p(a_{t+1} | h_t,\theta)$, we use a latent-variable model instead of directly using a neural network with a GAN-based loss as in previous works \citep{ho2016gail,bhattacharyya2018ps-gail}. As a result, the log-probability of the action is approximated by the Evidence Lower Bound (ELBO) using a posterior distribution $q_\phi (z |a_{t+1},h_t)$ parameterized by $\phi$

\begin{equation}
    ELBO(\theta, \phi) = E_q \log p(a_{t+1}|z, h_t) - KL(q(z|h_t,a_{t+1})||p(z|h_t))
\end{equation}

This is similar to the loss form of conditional Variational Auto-Encoder (cVAE) \citep{sohn2015learning}. Note that we can output a new observation $o_{t+1}$ when the action $a_{t+1}$ is known. This leads to $q(z | h_t,a_{t+1}) = q(z | h_{t+1})$. Combined with Eq. \ref{likelihood_tau}, the formula of $ELBO$ over all time steps is

\begin{equation}
    \label{elbo_tau}
    ELBO_\tau(\theta, \phi) = \sum_{t=0}^{T-1} E_{z \sim q(z)} \log p(a_{t+1}|z, h_t) - \sum_{t=0}^{T-1} KL(q(z|h_{t+1})||p(z|h_t))
\end{equation}

where $p(z | h_0) = p(z) = \mathcal{N}(0, I)$ is the prior distribution of the latent variable $z$. We notice that using a neural network for encoding each $h_t$ to compute KL-divergence in Eq. \ref{elbo_tau} is expensive. When we minimize the KL-divergence term, we aim to encode all actions and observations of the trajectory $\tau$ into the same distribution of $z$. This means that we model the entire trajectory $\tau$ by a latent variable $z$. To achieve this goal, inspired by the Latent-ODE model \citep{rubanova2019lode}, we assume that the trajectory $\tau$ is generated from an Ordinary Differential Equation (ODE) in the latent space. We use an ODE-RNN network which runs backward in time to estimate the posterior distribution of $z$  given the entire history data $h_T$. Therefore, we can replace the encoder term, $KL$, in Eq. \ref{elbo_tau} simply by 

\begin{equation}
    \label{KL_term}
    \sum_{t=0}^{T-1} KL(q(z|h_t,a_{t+1})||p(z|h_t)) \rightarrow KL(q(z | h_T) || p(z))
\end{equation}

where $q(z | h_T) = \mathcal{N}(\mu_z, \sigma_z)$, and $(\mu_z, \sigma_z) = \textit{ODE-RNN}(h_T)$.

After estimating the latent distribution, we sample $z_1$ as a starting point from this distribution to generate the trajectory of actions. Given $z_1$, we use a neural ODE to represent the trajectory of actions in the latent space. We denote $z_t$ as the hidden state of the action $a_t$ at each time step $t$. $z_t$ can be estimated by solving the ODE with the starting point $z_1$ as follows

\begin{equation}
    \frac{dz}{dt} = f_\lambda(z, t); \quad (z_1,z_2,...,z_{T+k}) = ODESolve(f_\lambda, z_1, (1,...,T+k))
\end{equation}

Finally, each state $z_t$ at time step $t$ is used to predict the distribution of actions $p(a_t |z, h_{t-1})$ in Eq. \ref{elbo_tau}. 

\begin{equation}
    \label{rec_term}
    p(a_t|z, h_{t-1}) = p(a_t|z) = \mathcal{N}(\mu_{z_t}, \sigma_{z_t}); \quad (\mu_{z_t}, \sigma_{z_t}) = h(z_t)     
\end{equation}
where $h(.)$ is a Multi-Layer Perceptron (MLP). 

\begin{figure}[t]
\begin{center}
\includegraphics[scale=0.4]{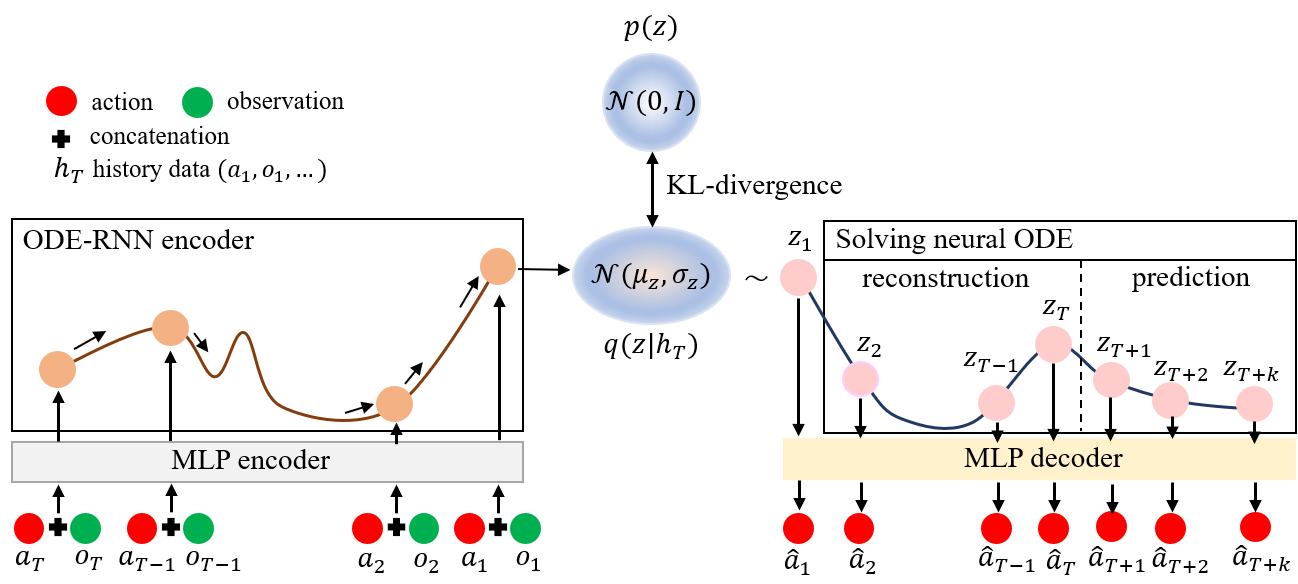}
\end{center}
\caption{Illustration of our proposed architecture for the conditional Latent ODEs (cLODE).}
\label{cLODE_architecture}
\vspace{-0.3cm}
\end{figure}

\subsection{Network architecture}

Fig. \ref{cLODE_architecture} illustrates the overview of our proposed architecture cLODE. Similar to the VAE-based generative models \citep{kingma2013auto,rezende2014stochastic}, the architecture includes two parts: an encoder and a decoder. The encoder is used to embed the history data into the distribution of the latent variable $z$. The decoder then generates the action based on the sampled hidden state. We describe the detail of each part in Appendix.




%% file: sections/40_results.tex
\section{Results}  \label{results}

\subsection{Experimental Setup}
We validate the performance of cLODE on the US 101 human driving data from the Next Generation SIMulation (NGSIM) dataset \citep{colyar2007us}. As a baseline, we use an imitative algorithm for autonomous driving, Adaptable Generative Prediction Networks for Autonomous Driving (AGen) \citep{si2019agen}. We can make a direct comparison between the two algorithms because both are based on the NGSIM dataset, which is a widely used benchmark dataset for many autonomous driving problems. 

The dimension of one observation pair is on $\mathbf{R}^{66}$, where we consider multiple concatenations of the observation history for the encoder network of cLODE. The number of vehicle agents throughout the whole experiment is fixed to 22 to incorporate multi-agent interactions into the driving data, although NGSIM also supports single-agent simulations. For training the 22 cLODE agents on the NGSIM data, we set batch size $B$, learning rate $l_r$, simulation update rate $dt$ as $B=50$, $l_r=1e-3$, $dt=0.1$, respectively.

\subsection{Evaluation Metrics}

For fair comparison, we adopt the same metric as AGen, which is root mean squared error (RMSE) of predicted vehicle roll-out trajectory with repect to the dataset Ground-truth. Specifically, if we want to evaluate RMSE of a value $v$ at time $t$ over $m$ samples,

\begin{equation}
\resizebox{140pt}{!}{%
${\rm RMSE } = \sqrt{\frac{1}{m} \sum^m_{i=1} \left( v^{(i)}_t - \hat{v}^{(i)}_t\right)^2}$}
\end{equation} 
where $v^{(i)}_t$ is the Ground-truth value of $v$ and $\hat{v}^{(i)}_t$ is the predicted value. Note that $i$ is the sample index. Also, if we define the vehicle positional error at time $t$ of the top-view trajectory as $p_t$, with corresponding longitudinal error $x_t$ and lateral error $y_t$: $p_t = \sqrt{x_t^2 + y_t^2}$. Note that both metrics are calculated along the base coordinate which is fixed at the start point of the vehicle trajectory. As such, we can calculate the total positional RMSE($p_t$) as well as the longitudinal RMSE($x_t$) and the lateral RMSE($y_t$) if trajectory data are given.

\subsection{Prediction Results}

\subsubsection{Computation time}
All experiments are conducted with an ordinary desktop computer equipped with 3.6GHz Intel Core i7 CPU, 64GB RAM and NVIDIA GeForce 1080 GPU. The memory consumption of cLODE at test time on GPU is up to 598MB, which is way lower than that of AGen since it takes 6,547MB memory on the same GPU. This indicates that cLODE provides more memory-efficient inference than the baseline. However, cLODE needs longer time to roll-out the predicted vehicle trajectories, as it requires 71.18(s) to predict 200 time steps where it took 57.79(s) for AGen. This result is equivalent to 0.36(s) per time step for cLODE versus 0.29(s) per time step for AGen, meaning that cLODE makes 1.24 times slower prediction. We conjecture this is mainly because cLODE pays constant memory cost along the neural ODE computation. Nevertheless, such a downside is deemed to be a marginal sacrifice considering about 10-fold gain on the memory consumption efficiency.

\subsubsection{RMSE}

\begin{table}[!t]
\begin{minipage}{.46\textwidth}
\caption{Positional RMSEs at 2.5(s) with 22 vehicle agents.}
\label{table:rmse}
\centering
\resizebox{7cm}{!}{%
    \begin{tabular}{m{9em} | m{4.5em} m{2.5em} m{2em}}
        \toprule
         & Longitudinal (m)     & Lateral (m) & Total (m) \\
        \midrule
        AGen & 1.64 & 1.89 &  2.51    \\
        cLODE with 5 obs. & 1.44 & 1.64 & 2.18     \\
        cLODE with 10 obs. & 1.47 & 1.70 & 2.25      \\
        cLODE with 20 obs. & 1.45 & 1.70 & 2.24      \\
        cLODE with 50 obs. & 1.53 & 1.79 & 2.36      \\
        cLODE with 100 obs. & 1.50 & 1.74 & 2.29     \\
    \bottomrule
    \end{tabular}
}
\end{minipage}
\hfill
\begin{minipage}{.48\textwidth}
\centering
  \includegraphics[scale=0.4]{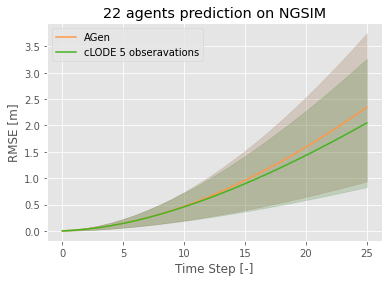}
\captionof{figure}{Positional RMSE comparison between AGen and cLODE with standard deviation.}
\label{fig:agenclode25}
\end{minipage}
\vspace{-0.5cm}
\end{table}




\begin{table}[!t]
\begin{minipage}{.46\textwidth}
\centering
  \includegraphics[scale=0.4]{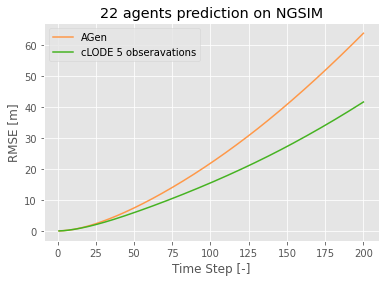}
\captionof{figure}{Positional RMSE comparison between AGen and cLODE up to 200 time stpes.}
\label{fig:agenclode200}
\end{minipage}
\hfill
\begin{minipage}{.48\textwidth}
\centering
  \includegraphics[scale=0.4]{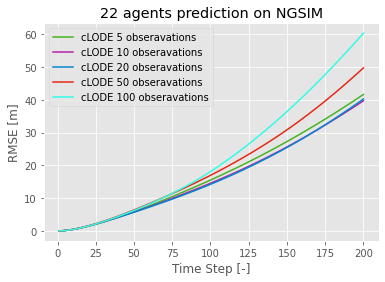}
\captionof{figure}{Ablation study of observation history length on RMSE performance.}
\label{fig:clodeabl}
\end{minipage}
\vspace{-0.5cm}
\end{table}



The positional RMSE results at 2.5(s) on the vehicle trajectory are shown in Table \ref{table:rmse}. We focus on the record at this specific time because as time step increases the vehicle position predictions also tend to increasingly deviate from the Ground-truth data in general. As indicated in the table, even though AGen is an domain-adapted model boosted by a generative predictor trained with PS-GAIL, cLODE outperformed over all RMSE metrics. The comparison between the RMSE results of the two models up to 2.5(s) is delineated in Fig. \ref{fig:agenclode25}. The figure is with standard deviation of the data over 330 trajectory samples and 22 vehicle agents, which shows that the difference in variance is not significant. The RMSE throughout 0$\sim$200 time steps is also depicted in Fig. \ref{fig:agenclode200}, which suggests that the RMSE is consistently smaller with cLODE model.

\subsubsection{Ablation Study on the Length of Observation History}


As the transition model of real-world vehicle dynamics is not a Markov Process, observation length of the vehicle state when we train the cLODE model may relate to prediction performance. With the data shown in Table \ref{table:rmse} and Fig. \ref{fig:clodeabl} we noted that even with just 5 observations, the cLODE model successfully predicts the Ground-truth vehicle trajectories, outperforming the AGen results. Surprisingly, stacking more observations in the history do not help increasing the performance. After augmenting to more than 20 observations, the trajectory prediction performance gradually deteriorates. We conjecture this is because, during training, the redundant information fed to the model negatively affects the learning process. This phenomenon is commonly known as causal confusion in imitation learning \citep{dehaan2019causal}.

%% file: sections/50_discussion_and_conclusions.tex
\section{Discussion and conclusions}  \label{discussion&conclusions}
In this report, we proposed a generative imitation learning model for autonomous driving, the conditional latent ordinary differential equation (cLODE). cLODE combines VAE with neural ODE for a more accurate prediction of multi-agent vehicle trajectories. Utilizing a Latent-ODE-inspired architecture, the model successfully predicts the vehicle state transitions in the latent space, outperforming the adaptive generative prediction networks AGen on RMSE metrics on myriads of NGSIM experiments. Furthermore, we found that the appropriate length of the observation history helps increase the prediction performance from the ablation studies.

%% file: sections/60_appendix.tex
\appendix

\section{Appendix}
\subsection{The Details of Network Architecture} 
\paragraph{Encoder} For each time step $t$, we first concatenate the observation $o_t$ and the action $a_t$, and then use a MLP with 6 fully connected layers to output an embedding vector $e_t$. All embedding vectors at all time steps are propagated backward into the first time step $t=1$ via an ODE-RNN model to estimate a representation of the observed trajectory $\tau$. ODE-RNN estimates an intermediate embedding $\hat{e}_{t-1}$ at the step $t-1$ based on $(e_t,...,e_T)$ by solving the ODE, and then updates the current embedding $e_t$ with $\hat{e}_t$ via a GRU. 

\paragraph{Decoder} The neural ODE used for decoding trajectories has 3 layers with 256 units for each. After estimating the hidden variable $z_t$ by solving the ODE, we use an MLP with 3 fully connected layers to predict the action $\hat{a}_t$.

\paragraph{Training and Prediction}  We train cLODE by optimizing the ELBO loss computed from Eq. \ref{elbo_tau},\ref{KL_term}, and \ref{rec_term}. After the model is trained, it can predict an action at any time given an arbitrary length of the history data. 

%% file: main.bbl
\begin{thebibliography}{22}
\makeatletter
\newcommand{\dinatlabel}[1]%
{\ifNAT@numbers\else\NAT@biblabelnum{#1}\hspace{2\labelsep}\fi}
\makeatother
\expandafter\ifx\csname natexlab\endcsname\relax\def\natexlab#1{#1}\fi
\expandafter\ifx\csname url\endcsname\relax\def\url#1{\texttt{#1}}\fi

\bibitem[Bain und Sammut(1995)]{bain1995bc}
\dinatlabel{Bain und Sammut 1995} \textsc{Bain}, Michael~; \textsc{Sammut},
  Claude:
\newblock A Framework for Behavioural Cloning.
\newblock In: \emph{Machine Intelligence 15}, 1995

\bibitem[Bhattacharyya u.\,a.(2018)Bhattacharyya, Phillips, Wulfe, Morton,
  Kuefler und Kochenderfer]{bhattacharyya2018ps-gail}
\dinatlabel{Bhattacharyya u.\,a. 2018} \textsc{Bhattacharyya}, Raunak~P.~;
  \textsc{Phillips}, Derek~J.~; \textsc{Wulfe}, Blake~; \textsc{Morton},
  Jeremy~; \textsc{Kuefler}, Alex~; \textsc{Kochenderfer}, Mykel~J.:
\newblock Multi-Agent Imitation Learning for Driving Simulation.
\newblock In: \emph{2018 IEEE/RSJ International Conference on Intelligent
  Robots and Systems (IROS)}, 2018, S.~1534--1539

\bibitem[Che u.\,a.(2018)Che, Purushotham, Cho, Sontag und
  Liu]{che2018rnndecay}
\dinatlabel{Che u.\,a. 2018} \textsc{Che}, Zhengping~; \textsc{Purushotham},
  Sanjay~; \textsc{Cho}, Kyunghyun~; \textsc{Sontag}, David~; \textsc{Liu},
  Yan:
\newblock Recurrent Neural Networks for Multivariate Time Series with Missing
  Values.
\newblock In: \emph{Scientific Reports}
\newblock 8 (2018), 04

\bibitem[Chen u.\,a.(2018)Chen, Rubanova, Bettencourt und
  Duvenaud]{chen2018neuralode}
\dinatlabel{Chen u.\,a. 2018} \textsc{Chen}, Ricky T.~Q.~; \textsc{Rubanova},
  Yulia~; \textsc{Bettencourt}, Jesse~; \textsc{Duvenaud}, David~K.:
\newblock Neural Ordinary Differential Equations.
\newblock In: \textsc{Bengio}, S. (Hrsg.)~; \textsc{Wallach}, H. (Hrsg.)~;
  \textsc{Larochelle}, H. (Hrsg.)~; \textsc{Grauman}, K. (Hrsg.)~;
  \textsc{Cesa-Bianchi}, N. (Hrsg.)~; \textsc{Garnett}, R. (Hrsg.):
  \emph{Advances in Neural Information Processing Systems} Bd.~31, Curran
  Associates, Inc., 2018. --
\newblock URL
  \url{https://proceedings.neurips.cc/paper/2018/file/69386f6bb1dfed68692a24c8686939b9-Paper.pdf}

\bibitem[Colyar und Halkias(2007)]{colyar2007us}
\dinatlabel{Colyar und Halkias 2007} \textsc{Colyar}, James~; \textsc{Halkias},
  John:
\newblock US highway 101 dataset.
\newblock In: \emph{Federal Highway Administration (FHWA), Tech. Rep}
\newblock FHWA-HRT07-030 (2007)

\bibitem[Finn u.\,a.(2016)Finn, Levine und Abbeel]{finn2016irl}
\dinatlabel{Finn u.\,a. 2016} \textsc{Finn}, Chelsea~; \textsc{Levine},
  Sergey~; \textsc{Abbeel}, Pieter:
\newblock Guided Cost Learning: Deep Inverse Optimal Control via Policy
  Optimization.
\newblock In: \emph{Proceedings of the 33rd International Conference on
  International Conference on Machine Learning - Volume 48}, JMLR.org, 2016
\newblock (ICML'16), S.~49–58

\bibitem[de~Haan u.\,a.(2019)de~Haan, Jayaraman und Levine]{dehaan2019causal}
\dinatlabel{de~Haan u.\,a. 2019} \textsc{Haan}, Pim de~; \textsc{Jayaraman},
  Dinesh~; \textsc{Levine}, Sergey:
\newblock Causal Confusion in Imitation Learning.
\newblock In: \emph{arXiv}
\newblock cs.LG 1905.11979 (2019)

\bibitem[Ho und Ermon(2016)]{ho2016gail}
\dinatlabel{Ho und Ermon 2016} \textsc{Ho}, Jonathan~; \textsc{Ermon}, Stefano:
\newblock Generative Adversarial Imitation Learning.
\newblock In: \textsc{Lee}, D. (Hrsg.)~; \textsc{Sugiyama}, M. (Hrsg.)~;
  \textsc{Luxburg}, U. (Hrsg.)~; \textsc{Guyon}, I. (Hrsg.)~; \textsc{Garnett},
  R. (Hrsg.): \emph{Advances in Neural Information Processing Systems} Bd.~29,
  Curran Associates, Inc., 2016. --
\newblock URL
  \url{https://proceedings.neurips.cc/paper/2016/file/cc7e2b878868cbae992d1fb743995d8f-Paper.pdf}

\bibitem[Ho u.\,a.(2016)Ho, Gupta und Ermon]{ho2016irl}
\dinatlabel{Ho u.\,a. 2016} \textsc{Ho}, Jonathan~; \textsc{Gupta}, Jayesh~K.~;
  \textsc{Ermon}, Stefano:
\newblock Model-Free Imitation Learning with Policy Optimization.
\newblock In: \emph{Proceedings of the 33rd International Conference on
  International Conference on Machine Learning - Volume 48}, JMLR.org, 2016
\newblock (ICML'16), S.~2760–2769

\bibitem[Hussein u.\,a.(2017)Hussein, Gaber, Elyan und
  Jayne]{hussein2017reviewimitation}
\dinatlabel{Hussein u.\,a. 2017} \textsc{Hussein}, Ahmed~; \textsc{Gaber},
  Mohamed~M.~; \textsc{Elyan}, Eyad~; \textsc{Jayne}, Chrisina:
\newblock Imitation Learning: A Survey of Learning Methods.
\newblock In: \emph{ACM Comput. Surv.}
\newblock 50 (2017), apr, Nr.~2. --
\newblock URL \url{https://doi.org/10.1145/3054912}. --
\newblock ISSN 0360-0300

\bibitem[Jena u.\,a.(2021)Jena, Liu und Sycara]{jena2021augmentinggail}
\dinatlabel{Jena u.\,a. 2021} \textsc{Jena}, Rohit~; \textsc{Liu}, Changliu~;
  \textsc{Sycara}, Katia:
\newblock Augmenting GAIL with BC for sample efficient imitation learning.
\newblock In: \textsc{Kober}, Jens (Hrsg.)~; \textsc{Ramos}, Fabio (Hrsg.)~;
  \textsc{Tomlin}, Claire (Hrsg.): \emph{Proceedings of the 2020 Conference on
  Robot Learning} Bd.~155, PMLR, 16--18 Nov 2021, S.~80--90. --
\newblock URL \url{https://proceedings.mlr.press/v155/jena21a.html}

\bibitem[Kingma und Welling(2013)]{kingma2013auto}
\dinatlabel{Kingma und Welling 2013} \textsc{Kingma}, Diederik~P.~;
  \textsc{Welling}, Max:
\newblock Auto-encoding variational bayes.
\newblock In: \emph{arXiv preprint arXiv:1312.6114}
\newblock (2013)

\bibitem[Kuefler u.\,a.(2017)Kuefler, Morton, Wheeler und
  Kochenderfer]{kuefler2017gail-driving}
\dinatlabel{Kuefler u.\,a. 2017} \textsc{Kuefler}, Alex~; \textsc{Morton},
  Jeremy~; \textsc{Wheeler}, Tim~; \textsc{Kochenderfer}, Mykel:
\newblock Imitating driver behavior with generative adversarial networks.
\newblock In: \emph{2017 IEEE Intelligent Vehicles Symposium (IV)}, 2017,
  S.~204--211

\bibitem[Lipton u.\,a.(2016)Lipton, Kale und Wetzel]{lipton2016rnndecay}
\dinatlabel{Lipton u.\,a. 2016} \textsc{Lipton}, Zachary~C.~; \textsc{Kale},
  David~C.~; \textsc{Wetzel}, Randall:
\newblock \emph{Modeling Missing Data in Clinical Time Series with RNNs}.
\newblock 2016. --
\newblock URL \url{https://arxiv.org/abs/1606.04130}

\bibitem[Rezende u.\,a.(2014)Rezende, Mohamed und
  Wierstra]{rezende2014stochastic}
\dinatlabel{Rezende u.\,a. 2014} \textsc{Rezende}, Danilo~J.~;
  \textsc{Mohamed}, Shakir~; \textsc{Wierstra}, Daan:
\newblock Stochastic backpropagation and approximate inference in deep
  generative models.
\newblock In: \emph{International conference on machine learning}
\newblock PMLR (Veranst.), 2014, S.~1278--1286

\bibitem[Rubanova u.\,a.(2019)Rubanova, Chen und Duvenaud]{rubanova2019lode}
\dinatlabel{Rubanova u.\,a. 2019} \textsc{Rubanova}, Yulia~; \textsc{Chen},
  Ricky T.~Q.~; \textsc{Duvenaud}, David~K.:
\newblock Latent Ordinary Differential Equations for Irregularly-Sampled Time
  Series.
\newblock In: \textsc{Wallach}, H. (Hrsg.)~; \textsc{Larochelle}, H. (Hrsg.)~;
  \textsc{Beygelzimer}, A. (Hrsg.)~; \textsc{Alch\'{e}-Buc}, F.
  d\textquotesingle (Hrsg.)~; \textsc{Fox}, E. (Hrsg.)~; \textsc{Garnett}, R.
  (Hrsg.): \emph{Advances in Neural Information Processing Systems} Bd.~32,
  Curran Associates, Inc., 2019, S.~NA. --
\newblock URL
  \url{https://proceedings.neurips.cc/paper/2019/file/42a6845a557bef704ad8ac9cb4461d43-Paper.pdf}

\bibitem[Sasaki u.\,a.(2020)Sasaki, Yohira und Kawaguchi]{sasaki2020abc}
\dinatlabel{Sasaki u.\,a. 2020} \textsc{Sasaki}, Fumihiro~; \textsc{Yohira},
  Tetsuya~; \textsc{Kawaguchi}, Atsuo:
\newblock Adversarial behavioral cloning.
\newblock In: \emph{Advanced Robotics}
\newblock 34 (2020), Nr.~9, S.~592--598. --
\newblock URL \url{https://doi.org/10.1080/01691864.2020.1729237}

\bibitem[Si u.\,a.(2019)Si, Wei und Liu]{si2019agen}
\dinatlabel{Si u.\,a. 2019} \textsc{Si}, Wenwen~; \textsc{Wei}, Tianhao~;
  \textsc{Liu}, Changliu:
\newblock AGen: Adaptable Generative Prediction Networks for Autonomous
  Driving.
\newblock In: \emph{2019 IEEE Intelligent Vehicles Symposium (IV)}, 2019,
  S.~281--286

\bibitem[Sohn u.\,a.(2015)Sohn, Lee und Yan]{sohn2015learning}
\dinatlabel{Sohn u.\,a. 2015} \textsc{Sohn}, Kihyuk~; \textsc{Lee}, Honglak~;
  \textsc{Yan}, Xinchen:
\newblock Learning structured output representation using deep conditional
  generative models.
\newblock In: \emph{Advances in neural information processing systems}
\newblock 28 (2015)

\bibitem[Song u.\,a.(2018)Song, Ren, Sadigh und Ermon]{song2018ma-gail}
\dinatlabel{Song u.\,a. 2018} \textsc{Song}, Jiaming~; \textsc{Ren}, Hongyu~;
  \textsc{Sadigh}, Dorsa~; \textsc{Ermon}, Stefano:
\newblock Multi-Agent Generative Adversarial Imitation Learning.
\newblock In: \textsc{Bengio}, S. (Hrsg.)~; \textsc{Wallach}, H. (Hrsg.)~;
  \textsc{Larochelle}, H. (Hrsg.)~; \textsc{Grauman}, K. (Hrsg.)~;
  \textsc{Cesa-Bianchi}, N. (Hrsg.)~; \textsc{Garnett}, R. (Hrsg.):
  \emph{Advances in Neural Information Processing Systems} Bd.~31, Curran
  Associates, Inc., 2018. --
\newblock URL
  \url{https://proceedings.neurips.cc/paper/2018/file/240c945bb72980130446fc2b40fbb8e0-Paper.pdf}

\bibitem[Torabi u.\,a.(2018)Torabi, Warnell und Stone]{torabi2018bco}
\dinatlabel{Torabi u.\,a. 2018} \textsc{Torabi}, Faraz~; \textsc{Warnell},
  Garrett~; \textsc{Stone}, Peter:
\newblock Behavioral Cloning from Observation.
\newblock In: \emph{Proceedings of the 27th International Joint Conference on
  Artificial Intelligence}, AAAI Press, 2018
\newblock (IJCAI'18), S.~4950–4957. --
\newblock ISBN 9780999241127

\bibitem[Wang u.\,a.(2019)Wang, Adeli, Chiu, Huang und Niebles]{wang2019ilhpp}
\dinatlabel{Wang u.\,a. 2019} \textsc{Wang}, B.~; \textsc{Adeli}, E.~;
  \textsc{Chiu}, H.~; \textsc{Huang}, D.~; \textsc{Niebles}, J.:
\newblock Imitation Learning for Human Pose Prediction.
\newblock In: \emph{2019 IEEE/CVF International Conference on Computer Vision
  (ICCV)}.
\newblock Los Alamitos, CA, USA~: IEEE Computer Society, nov 2019,
  S.~7123--7132. --
\newblock URL \url{https://doi.ieeecomputersociety.org/10.1109/ICCV.2019.00722}

\end{thebibliography}
